\documentclass[sigconf]{acmart}
\AtBeginDocument{%
  \providecommand\BibTeX{{%
    \normalfont B\kern-0.5em{\scshape i\kern-0.25em b}\kern-0.8em\TeX}}}

\setcopyright{acmcopyright}
\copyrightyear{2024}
\acmYear{2024}
\acmDOI{XXXXXXX.XXXXXXX}

\acmConference[WSDM '24]{the 17th ACM International Conference on Web Search and Data Mining}{March 04--08, 2024}{Mérida, Mexico}
\acmPrice{15.00}
\acmISBN{978-1-4503-XXXX-X/18/06}

\usepackage{multirow}
\usepackage{subfigure}
\usepackage{amsmath}
\newtheorem{theorem}{Theorem}
\usepackage{amsthm}
\usepackage{booktabs}
\usepackage[switch]{lineno}
\usepackage{array}
\usepackage{arydshln}
\usepackage{bm}
\usepackage{algorithm}
\usepackage{algorithmic}
\newcommand{\code}[0]{\url{https://github.com/ZzoomD/FairGKD/}}

\begin{document}

\title{The Devil is in the Data: Learning Fair Graph Neural Networks \\ via Partial Knowledge Distillation}

\author{Yuchang Zhu}
\affiliation{%
  \institution{Sun Yat-sen University}
  \country{}
}
\email{zhuych27@mail2.sysu.edu.cn}

\author{Jintang Li}
\affiliation{%
  \institution{Sun Yat-sen University}
  \country{}
  }
\email{lijt55@mail2.sysu.edu.cn}

\author{Liang Chen}
\authornote{Corresponding author.}
\affiliation{%
  \institution{Sun Yat-sen University}
  \country{}
}
\email{chenliang6@mail.sysu.edu.cn}

\author{Zibin Zheng}
\affiliation{%
 \institution{Sun Yat-sen University}
 \country{}
 }
 \email{zhzibin@mail.sysu.edu.cn}


\begin{abstract}
  Graph neural networks (GNNs) are being increasingly used in many high-stakes tasks, and as a result, there is growing attention on their fairness recently. GNNs have been shown to be unfair as they tend to make discriminatory decisions toward certain demographic groups, divided by sensitive attributes such as gender and race. While recent works have been devoted to improving their fairness performance, they often require accessible demographic information. This greatly limits their applicability in real-world scenarios due to legal restrictions. To address this problem, we present a demographic-agnostic method to learn fair GNNs via knowledge distillation, namely \textbf{FairGKD}. Our work is motivated by the empirical observation that training GNNs on partial data (i.e., only node attributes or topology data) can improve their fairness, albeit at the cost of utility. To make a balanced trade-off between fairness and utility performance, we employ a set of fairness experts (i.e., GNNs trained on different partial data) to construct the synthetic teacher, which distills fairer and informative knowledge to guide the learning of the GNN student. 
  Experiments on several benchmark datasets demonstrate that FairGKD, which does not require access to demographic information, significantly improves the fairness of GNNs by a large margin while maintaining their utility.\footnote{Our code is available via: \code.}
\end{abstract}





\maketitle

\section{Introduction}
 Graph neural networks (GNNs) have demonstrated superior performance in various applications~\cite{fan2019graph,aykent2022gbpnet,jiang2022graph}. However, the increasing application of GNNs in high-stakes tasks, such as credit scoring~\cite{wang2021temporal} and fraud detection~\cite{liu2020alleviating}, has raised concerns regarding their fairness, as highlighted by recent works~\cite{rahman2019fairwalk,dong2022edits}. A widely accepted view is that the source of biases that result in the fairness problem of GNNs is the training data~\cite{beutel2017data,dai2021say,dong2022edits}. GNNs inherit or even amplify these biases through message passing~\cite{dai2021say}, leading to biased decision-making toward certain demographic groups divided by sensitive attributes such as race, gender, etc. As a result, such discriminatory decisions may potentially cause economic problems and even social problems~\cite{mukerjee2002multi,mehrabi2021survey}.



 Over the past few years, efforts~\cite{dai2021say,jiang2022fmp,spinelli2021fairdrop} have been made to improve the fairness performance of GNNs. A well-studied approach is to mitigate fairness-related biases by modifying the training data, such as reducing the connection of nodes within the same demographic group~\cite{spinelli2021fairdrop} or preprocessing data to minimize the distribution distance between demographic groups~\cite{dong2022edits}. In this regard, the trained GNNs inherit less bias from unbiased training data. Furthermore, another popular approach is to address the fairness problem from a training perspective such as adversarial learning, which aims to learn a fair GNN model to generate node embedding independent of the sensitive attribute~\cite{bose2019compositional,dai2021say}. 
 Despite significant progress, prior works often assume that sensitive attributes (i.e., demographic information), are fully or partially accessible. However, due to privacy and legal restrictions~\cite{chai2022fairness,hashimoto2018fairness} on sensitive attributes, such an assumption may not always hold in real-world scenarios.
 Although recent advances~\cite{zhu2022learning,lahoti2020fairness} have explored improving fairness without demographic information for independent and identically distributed (IID) data, these works cannot be directly applied to graph data due to complicated relations between instances.
 To this end, a natural question arises:{\itshape How can we learn fair GNNs without demographic information?}

\begin{figure}[!t]
  \centering
  \includegraphics[width=\linewidth]{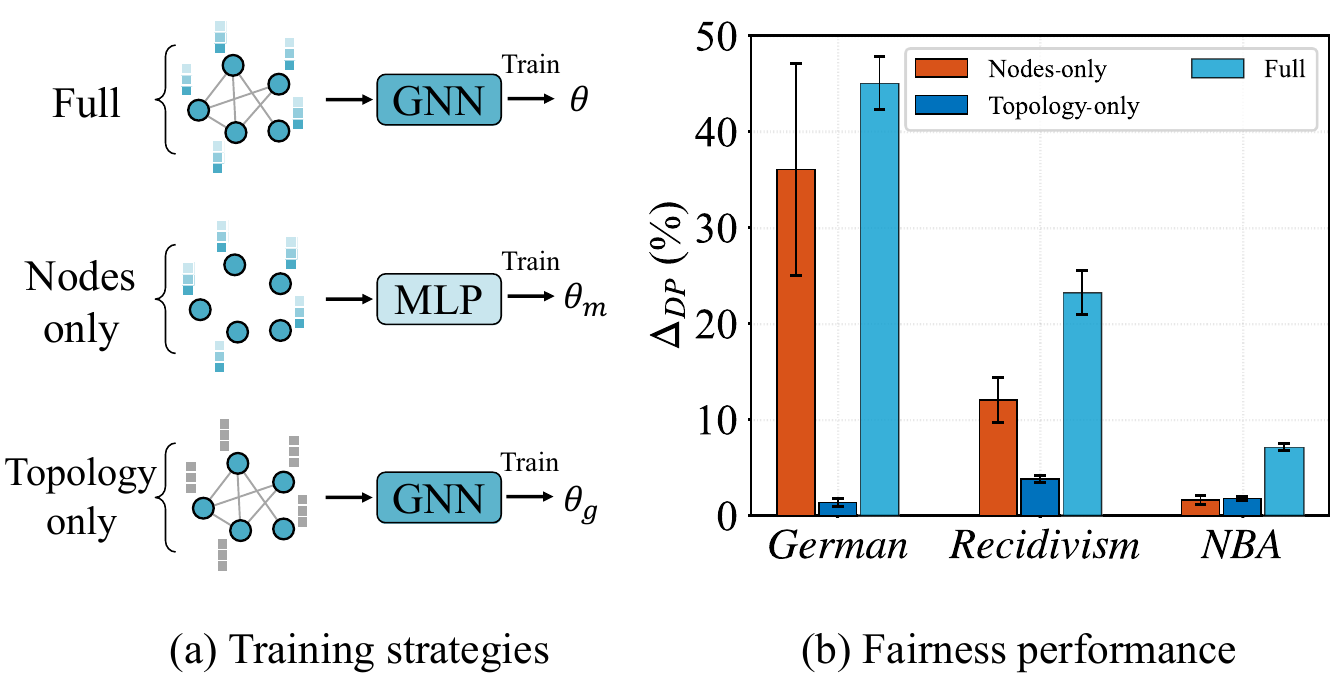}
  \caption{Fairness performance comparison of different training strategies. A smaller $\Delta_{DP}$ means better fairness performance. Partial data (i.e., nodes-only or topology-only) training improves fairness performance.}
  \label{fig:intro}
\end{figure}

We find that previous efforts focus on mitigating group-level biases, i.e., biases defined by the difference between demographic groups, resulting in the requirement for accessing the sensitive attribute. For example, FairDrop~\cite{spinelli2021fairdrop} reduces the connections between intra-group nodes, i.e., nodes within the same demographic group. Here, ``intra-group'' implicitly means bias mitigation at the group level. As such, prior approaches are highly dependent on the accessible sensitive attribute. In the preliminary study of Section \ref{sec:preliminary}, we find that mitigating the higher-level biases, i.e., bias in node attributes or the graph topology, also improves fairness but without accessing the sensitive attribute. Here, higher-level biases emphasize a wider range of biases than group-level biases, and in this paper refer to biases in node attributes or graph topology. Specifically, considering that the trained model inherits biases from the training data~\cite{dai2021say,ling2023learning}, using a portion of the data for training may be a straightforward solution to mitigate the higher-level biases. Thus, we compare fairness performance across three model training strategies, i.e., using only graph topology, only node attributes, and full graph data. The first two training strategies are referred to as {\itshape partial data training}, representing mitigating biases in node attributes, and graph topology, respectively. As shown in Figure \ref{fig:intro}, our comparison demonstrates that {\itshape partial data training improves fairness performance but inevitably sacrifices utility.} This observation indicates that {\itshape mitigating higher-level biases also improves fairness but without accessing the sensitive attribute.} 


In light of our observation, we propose the utilization of partial data training as a means to alleviate higher-level biases, thereby enhancing fairness in a demographic-agnostic manner. However, the suboptimal utility performance associated with partial data training renders it impractical for training inference GNNs using such an approach. Drawing inspiration from the notable success of knowledge distillation~\cite{hinton2015distilling}, employing partial data training to optimize fairer teacher models for guiding the learning process of GNN students emerges as a potentially effective solution. Thus, we propose a demographic-agnostic fairness method built upon partial data training and knowledge distillation paradigm, namely \textbf{FairGKD}. Specifically, FairGKD employs a set of fairness experts (i.e., models trained on partial data) to construct a synthetic teacher for distilling fair knowledge. Then, FairGKD learns fair GNN students with the guidance of fair knowledge. Additionally, FairGKD uses an adaptive algorithm to further achieve the trade-off between fairness and utility. Our contributions can be summarized as follows:
\begin{itemize}
\item We study a novel problem for learning fair GNNs without demographic information. To the best of our knowledge, our work is the first to explore fairness without demographic information on graph-structured data.

\item We propose FairGKD, a simple yet effective method that learns fair GNNs with the guidance of a fair synthetic teacher to mitigate their higher-level biases.

\item We conduct experiments on several real-world datasets to verify the effectiveness of the proposed method in improving fairness while preserving utility. 
\end{itemize}

\section{Related Work}
\label{sec:related_work}

\subsection{Fairness in Graph}
Fairness in the graph attracts increasing attention due to the superior performance of GNNs in different scenarios~\cite{kipf2016semi,hamilton2017inductive,xu2018powerful}. Commonly used fairness notions can be summarized as group fairness, individual fairness, and counterfactual fairness~\cite{dwork2012fairness,2020InFoRM,gear}. In this paper, we focus on group fairness which highlights that the model neither favors nor harms any demographic groups defined by the sensitive attribute. Efforts to improve group fairness have incorporated mitigating biases in graph-structured data~\cite{current2022fairmod,laclau2021all} and constructing a fair GNNs learning framework~\cite{agarwal2021towards}. Mitigating data biases involves modifying the graph topology (i.e., adjacency matrix)~\cite{li2021dyadic} and node attributes~\cite{dong2022edits}, which provides clean data for training GNNs. For example, FairDrop~\cite{spinelli2021fairdrop} proposes an edge dropout algorithm to reduce connection between nodes within the same demographic group. EDITS~\cite{dong2022edits} modifies both graph topology and node attributes with an objective that minimizes distribution distance between several demographic groups. In the fair GNN learning framework construction, adversarial learning~\cite{NIPS2014_5ca3e9b1} is a commonly used approach to learning a fair GNN for generating node representations or making decisions independent of the sensitive attribute. For example, FairGNN~\cite{dai2021say} utilizes an adversary to learn a fair GNN classifier which makes predictions independent of the sensitive attribute. FairVGNN~\cite{wang2022improving} proposes mitigating the sensitive attribute leakage through generative adversarial debiasing. However, these two types of approaches make a strong assumption that the sensitive attribute is accessible, i.e., the sensitive attribute is known. Such an assumption may not hold in real-world scenarios due to legal restrictions. Although~\cite{dai2021say} studies fair GNNs in limited sensitive attributes, it still requires the sensitive attribute. 

Different from previous works, this work aims to learn fair GNNs without accessing sensitive attributes. This requires novel techniques to overcome the challenges of learning in the absence of sensitive attributes. In addition, our proposed method, FairGKD, employs knowledge distillation to learn fair GNNs. It should be noted that~\cite{dong2023reliant} has focused on addressing the fairness problem in GNN-based knowledge distillation frameworks by adding a learnable proxy of bias for the shallow student model. In contrast, FairGKD learns fair GNN students through a fairer teacher model, which is quite different from Dong et al. ~\cite{dong2023reliant}.

\subsection{Fairness without Demographics}
Since the legal and privacy limitations for accessing sensitive attributes, there are some progresses~\cite{10.1145/3340531.3411980,10.1145/3488560.3498493} to focus on fairness without demographic information in machine learning. For example, DRO~\cite{pmlr-v80-hashimoto18a} proposes a method based on distributionally robust optimization without access to demographics, which achieves fairness by improving the worst-case distribution. As the authors point out, DRO struggles to reduce the impact of noisy outliers, e.g., outliers from label noise. To avoid this impact, ARL~\cite{NEURIPS2020_07fc15c9} improves fairness by addressing computationally-identifiable errors and proposes adversarially reweighted learning to improve the utility of worst-case groups. Inspired by label smoothing improving fairness,~\cite{chai2022fairness} utilizes knowledge distillation to generate soft labels instead of label smoothing. Although prior works improve fairness without demographics, these works design algorithms based on IID data, and their effectiveness in non-IID data, e.g., graph data, remains unknown. Thus, this work focuses on improving the fairness of GNNs working in graph-structured data without accessing sensitive attributes. Different from ~\cite{chai2022fairness}, FairGKD aims to construct the fairer teacher and regards intermediate results~\cite{romero2014fitnets} as soft targets.

\section{Preliminaries}

\subsection{Notations}
For clarity in writing, consistent with prior works~\cite{dai2021say,wang2022improving}, we contextualize our proposed method and relevant proofs within the framework of a node classification task involving binary sensitive attributes and binary label settings. We represent an attributed graph by $\mathcal{G}=(\mathcal{V}, \mathcal{E}, \textbf{X})$ where $\mathcal{V}$ is a set of $\lvert \mathcal{V} \rvert = n$ nodes, $\mathcal{E}$ is a set of $\lvert \mathcal{E} \rvert = m$ edges. $\textbf{X} \in \mathbb{R}^{n\times d}$ is the node attribute matrix where $d$ is the node attribute dimension. $\overline{\textbf{X}} \in \mathbb{R}^{n\times d}$ is an all-one node attribute matrix. $S \in \lbrace0, 1\rbrace^{n}$ represents the binary sensitive attribute. $\tilde{\textbf{X}} \in \mathbb{R}^{n\times (d-1)}$ is the node attribute matrix without the sensitive attribute. $\textbf{A} \in \lbrace0, 1\rbrace^{n\times n}$ is the adjacency matrix. $\textbf{A}_{uv}=1$ represents that there exists edge $e_{uv} \in \mathcal{E}$ between the node $u$ and the node $v$, and $\textbf{A}_{uv}=0$ otherwise.  For node $u$ and node $v$, if $S_u=S_v$, these two nodes are within the same demographic group. GNNs update the node representation vector $h$ through aggregating messages of its neighbors. As such, existing GNNs consist of two steps: (1) message propagation and aggregation; (2) node representation updating. Thus, the $k$-th layer of GNNs can be defined as:
\begin{align}
\label{eq:gnn}
  a^{(k)}_v &= \operatorname{AGGREGATE}^{(k)}(\{h^{(k-1)}_u:u\in \mathcal{N}(v)\}),\\
  h^{(k)}_v &= \operatorname{UPDATE}^{(k)}(h^{(k-1)}_v,a^{(k)}_v),
\end{align}
where $\operatorname{AGGREGATE}^{(k)}(\cdot)$ and $\operatorname{UPDATE}^{(k)}(\cdot)$ represent aggregation function and update function in $k$-th layer, respectively. $\mathcal{N}(v)$ represents the set of nodes adjacent to node $v$. 


\subsection{Fairness Metrics}
\label{sec:metrics}

We focus on group fairness which highlights the outputs of the model are not biased against any demographic groups. For group fairness, {\itshape demographic parity}~\cite{dwork2012fairness} and {\itshape equal opportunity}~\cite{hardt2016equality} are two widely used evaluation metrics. In this paper, we utilize these two metrics to evaluate the fairness of models in the node classification task. Demographic parity (DP) requires that the prediction is independent of the sensitive attribute $S$ and equal opportunity (EO) requires the same true positive rate for each demographic group. Let $\hat{y}$ denote the node label prediction result of the classifier. $y \in \{0, 1\}$ denotes the node label ground truth. The DP and EO difference between the two demographic groups can be defined as:
\begin{align}
\label{eq:metric}
  \Delta_{DP} &= \lvert P(\hat{y}=1 \lvert S=0) - P(\hat{y}=1 \lvert S=1) \rvert, \\
  \Delta_{EO} &= \lvert P(\hat{y}=1 \lvert y=1,S=0) - P(\hat{y}=1 \lvert y=1,S=1) \rvert,
\end{align}
where small $\Delta_{DP}$ and $\Delta_{EO}$ imply fairer decision-making of GNNs.

\subsection{Problem Definition}
 This work aims to learn a fair GNN classifier $f_g(\cdot)$ which does not require accessing to demographics. With $\Delta_{DP}$ and $\Delta_{EO}$ as evaluation metrics, a fair GNN achieves minimum value for these two metrics. The problem of this paper can be formally defined as:
 
 \textbf{Problem Definition.} {\itshape Given a graph $\tilde{\mathcal{G}}=(\mathcal{V}, \mathcal{E}, \tilde{\textbf{X}})$, but non-accessible sensitive attributes, and partial node label $y$, learn a fair GNN $f_{g}$ for node classification task while maintaining utility.}
\begin{equation}
\label{eq:problem}
f_g(\tilde{\mathcal{G}},y) \rightarrow \hat{y}.
\end{equation}

\section{Impact of Partial Data Training}
\label{sec:preliminary}

Prior efforts~\cite{spinelli2021fairdrop,dai2021say,dong2022edits} to improve fairness have a strong assumption that demographic information is available. This is due to the fact that they aim to mitigate group-level biases, i.e., biases defined by the difference between demographic groups. Then, demographic-based data modifications or training strategies, such as reducing connections within intra-group nodes, are employed to mitigate such biases. As such, this results in the requirement for accessible demographic information. A natural question is raised: {\itshape Can we alleviate other biases while mitigating group-level biases?} 

\begin{figure}[!t]
  \centering
  \includegraphics[width=0.95\linewidth]{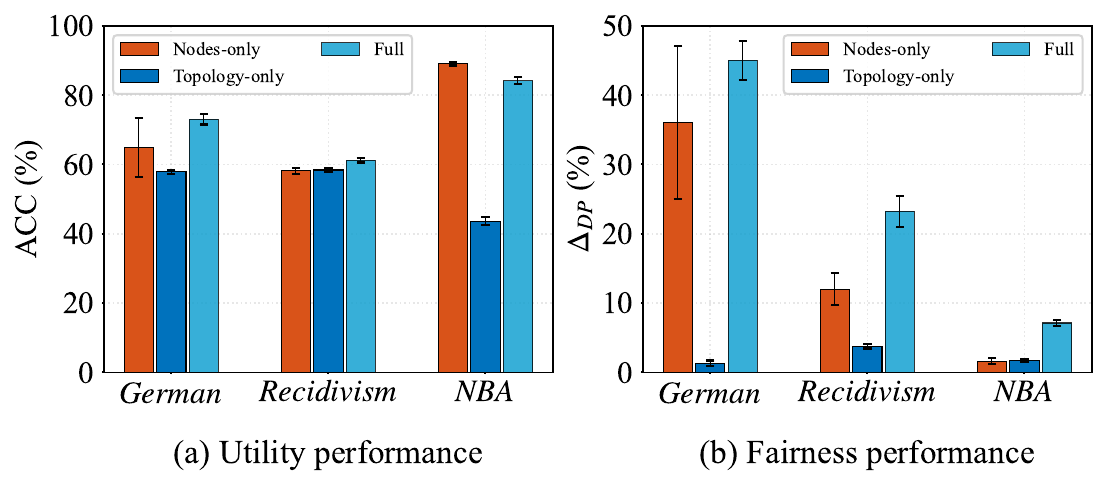}
  \caption{The preliminary results of three training strategies, i.e., full data, nodes-only, and topology-only. Partial data training improves fairness but sacrifices utility performance.}
  \label{fig:pre}
\end{figure}

Inspired by the fact that GNNs may inherit biases from training data~\cite{dai2021say,ling2023learning}, we speculate that training models on partial data, i.e., only node attributes or topology data, may be a natural solution. As shown in Figure~\ref{fig:pre}, we make a preliminary analysis for the fairness performance of different training strategies to verify our insight. We train GNNs on data with different components, which are described as follows: (1)~{\itshape Full data.} Train a 2-layer graph convolutional network (GCN)~\cite{kipf2016semi} classifier (i.e., a GCN layer followed by a linear layer) using the complete graph data with the binary cross-entropy (BCE) function as the loss function. (2)~{\itshape Nodes-only.} Train a 2-layer GCN classifier using only node attributes (i.e., the adjacent matrix is an identity matrix) with BCE as the loss function. This model can be regarded as a multi-layer perceptron (MLP). 
(3)~{\itshape Topology-only.} Train a 2-layer GCN classifier, which is the same as the classifier in {\itshape Full data}, using only graph topology (i.e., all-one node attributes matrix) with BCE as the loss function. Here, we refer to nodes-only and topology-only as partial data training.

\begin{figure*}[!htb]
  \centering
  \includegraphics[width=\textwidth]{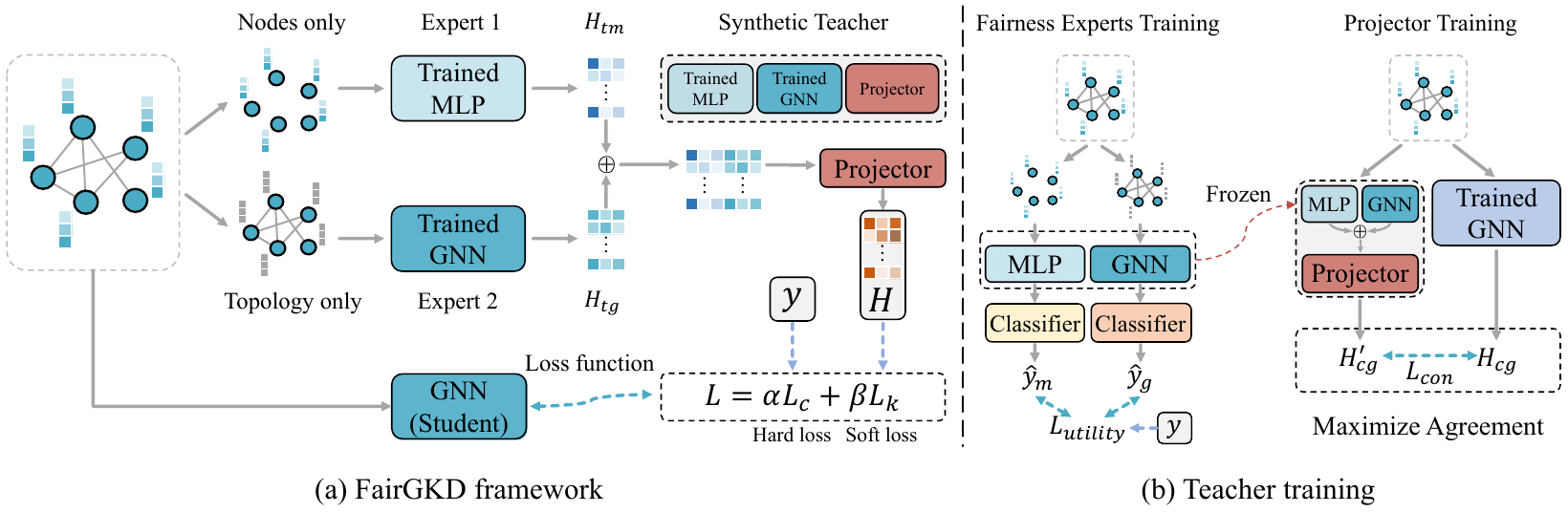}
  \caption{An overview of FairGKD framework. The synthetic teacher 
  distills fair and informative knowledge for guiding the learning of fair GNNs. ``Frozen'' means fixing model parameters.}
  \label{fig:overview}
\end{figure*}

We conduct the node classification experiment on three real-world datasets, i.e., {\itshape German}~\cite{agarwal2021towards}, {\itshape Recidivism}~\cite{agarwal2021towards}, and {\itshape NBA}~\cite{dai2021say}. We run this experiment 10 times to report results. All hyperparameters follow experimental settings in Section \ref{sec:exeps}.
The experimental results on all datasets are shown in Figure \ref{fig:pre}. We only show the accuracy (ACC) and $\Delta_{DP}$ performance due to the similar results on other metrics. From Figure \ref{fig:pre}, we make the following observations: 

\begin{itemize}
\item \textbf{Fairness.} Compared with full data training, partial data training (i.e., only node attributes or only graph topology) achieves a remarkable performance on fairness. 
\item \textbf{Utility.} Although partial data training performs better in fairness, it sacrifices utility performance.
\end{itemize}

The possible reason for the first observation is that: As the deep learning model inherits biases from training data~\cite{dai2021say,jiang2020identifying}, models trained on partial data only inherit biases in node attribute or graph topology, which results in better fairness performance. For example, graph topology in real-world scenarios exhibits the homophily of sensitive attributes, which is one of the sources of bias resulting in fairness problems~\cite{spinelli2021fairdrop,dong2023fairness}. {\itshape Topology-only} models are trained on such data to capture graph patterns but inevitably inherit bias from these patterns. In contrast, {\itshape Nodes-only} models avoid such bias by not taking into account graph topology. The same principle applies to node attributes. Here, partial data training makes trained models avoid inheriting biases in node attributes or graph topology, including fairness-relevant biases and fairness-irrelevant biases. Thus, we refer to all biases hidden in partial data (node attributes or graph topology) as higher-level biases which also encompass group-level biases. Mitigating group-level biases improves the fairness performance of GNNs~\cite{spinelli2021fairdrop,dong2022edits}. As such, alleviating higher-level biases also eliminates group-level biases, resulting in fairness performance improvement. 
For the second observation, there are two possible reasons to explain this: (1) GNNs are more powerful than MLP in representation and reasoning capacity, which is empirically proved by prior efforts~\cite{zhang2021graph,xu2018powerful}. (2) with partial data for training, the model misses part of the information in full data, which leads to the model utility sacrifice. 


Experimental results indicate that models trained on partial data present superior performance on fairness but sacrifice utility. To achieve the trade-off between fairness and utility, we propose FairGKD which is built upon knowledge distillation~\cite{hinton2015distilling} with partial data training, as shown in Section \ref{sec:method}.

\section{Methodology}
\label{sec:method}
Inspired by the preliminary analysis, we propose a method for learning fair GNNs, namely FairGKD. Here, we first give an overview of FairGKD. Then, we make a detailed description for each component of FairGKD, followed by the optimization objective. Finally, we present the theoretical proof and the complexity analysis in Appendix.

\subsection{Overview}
In this subsection, we provide an overview of FairGKD, which is illustrated in Figure \ref{fig:overview}. FairGKD is motivated by our empirical observation on partial data training. The goal of FairGKD is to learn a fair GNN without accessing sensitive attributes. To achieve this, several issues need to be tackled: (1) improving fairness without accessing the sensitive attribute; (2) avoiding utility sacrifice resulting from partial data training; (3) achieving a trade-off between fairness and utility. To overcome the first two challenges, FairGKD employs the partial data training strategy to mitigate higher-level biases. Then, following the knowledge distillation paradigm~\cite{fitnets,hinton2015distilling}, FairGKD employs two fairness experts (i.e., models trained on only node attributes or topology) to construct a synthetic teacher for distilling fair and informative knowledge. To overcome the third challenge, FairGKD utilizes an adaptive optimization algorithm to balance loss terms of fairness and utility. 

As shown in Figure \ref{fig:overview} (a), FairGKD consists of a synthetic teacher and a GNN student model denoted by $f_{t}$ and $f_{s}$, respectively. $f_{s}$ is a GNN classifier for the node classification task, mimicking the output of $f_{t}$. The synthetic teacher $f_{t}$ aims to distill fair and informative knowledge $H$ for the student model. Specifically, $f_{t}$ is comprised of two fairness experts, $f_{tm}$ and $f_{tg}$, and a projector $f_{tp}$. Here, $f_{tm}$ and $f_{tg}$, which are trained on only node attributes and only topology, alleviate higher-level biases without requiring access to sensitive attributes. Due to partial data training, $f_{tm}$ and $f_{tg}$ may generate fair yet uninformative node representations denoted by $H_{tm}$, $H_{tg}$. To bridge this gap, the projector $f_{tp}$ is used to combine these uninformative representations and performs mapping to generate informative representation $H$. $H$ will be regarded as additional supervision to assist the learning of $f_{s}$. Mimicking fair and informative representation $H$, $f_{s}$ tends to generate fair node representation while preserving utility. During each training epoch, $f_{s}$ takes full graph-structured data $\tilde{\mathcal{G}}=(\mathcal{V}, \mathcal{E}, \tilde{\textbf{X}})$ as input and predicts node labels. With a trained and fixed synthetic teacher as the teacher, $f_{s}$ is supervised by node label $y$ to maintain utility and node representation $H$ from the synthetic teacher to improve fairness. To further achieve the trade-off between fairness and utility for $f_{s}$, an adaptive optimization algorithm~\cite{zhao2022enhanced} is employed to balance the training influence between the utility loss term (hard loss) and the knowledge distillation loss term (soft loss). 

In summary, FairGKD is a demographic-agnostic framework that facilitates the learning of fair GNNs without sacrificing utility. Specifically, FairGKD is built upon knowledge distillation with partial data training. Different from the traditional teacher-student framework, FairGKD employs a synthetic teacher $f_{t}$, which is the combination of multiple models, to distill fair knowledge for learning fair GNNs. Benefiting from partial data training, FairGKD can learn fair GNNs without accessing sensitive attributes. Meanwhile, FairGKD can achieve fairness improvement for multiple sensitive attributes with just a single training session, as empirically demonstrated in Section~\ref{sec:exeps}.

\subsection{Synthetic Teacher}

In this subsection, we introduce the components of synthetic teacher $f_{t}$ and its optimization objective. The goal of $f_{t}$ is to distill fair and informative node representation $H$ for the student model. $f_{t}$ consists of two fairness experts $f_{tm}$, $f_{tg}$, and a projector $f_{tp}$. The MLP fairness expert $f_{tm}$ takes only node attributes without the sensitive attribute $\tilde{\textbf{X}}$ as input while outputting node representations $H_{tm}$. Meanwhile, the GNN fairness expert $f_{tg}$ takes graph $\overline{\mathcal{G}}=(\mathcal{V}, \mathcal{E}, \overline{\textbf{X}})$ with all-one node attributes matrix as input and outputs $H_{tg}$. Note that $\overline{\mathcal{G}}$ can be regarded as only graph topology due to the all-one node attributes matrix. The processing for obtaining the node representation $H_{tm}$ and $H_{tg}$ is as follows:
\begin{align}
\label{eq:mexperts_forward}
    H_{tm} &= f_{tm}(\tilde{\textbf{X}}), \\
\label{eq:gexperts_forward}
    H_{tg} &= f_{tg}(\overline{\mathcal{G}}).
\end{align}

$f_{tm}$ and $f_{tg}$ are trained on partial data, i.e., only node attributes and only topology, respectively. According to observation in Section \ref{sec:preliminary}, $f_{tm}$ and $f_{tg}$ only inherit bias in node attributes or topology. As a result, $H_{tm}$, $H_{tg}$ are fairer node representations due to mitigating higher-level biases. However, these two representations may be uninformative due to partial data training of $f_{tm}$ and $f_{tg}$, which can lead to poor performance in downstream tasks. With such representations as the additional supervised information, it is challenging to train a fair student model that preserves utility. To address this issue, FairGKD employs a projector $f_{tp}$ to merge these two representations to generate fair and informative representations $H$. Specifically, $f_{tp}$ takes the concatenation of $H_{tm}$ and $H_{tg}$ as input and outputs $H$. This can be summarized as:
\begin{equation}
\label{eq:t_forward}
H=f_{tp}(f_{tm}(\tilde{\textbf{X}}) \oplus f_{tg}(\overline{\mathcal{G}})),
\end{equation}
where $\oplus$ denotes the concatenate operation. Since partial data training mitigates higher-level biases,  $H_{tm}$ and $H_{tg}$ generated by the trained model display more fairness. To maintain such fairness, $f_{tp}$ aims to seek bias neutralization between $H_{tm}$ and $H_{tg}$, which results in fair representations $H$. Eq. (\ref{eq:t_forward}) combines the information from node attributes and graph topology while transforming $H_{tm}$ and $H_{tg}$ to a unified embedding space, similar to obtaining node representations with full data as input. As such, $f_{tp}$ can output informative yet fair representations $H$.



To ensure the implementation of the above design, FairGKD leverages a multi-step training scheme with a contrastive objective to train $f_{t}$. As shown in Figure \ref{fig:overview}(b), the multi-step training scheme can be divided into two steps, i.e., {\itshape fairness experts training} and {\itshape projector training}. (1) Fairness experts training aims to make fairness experts inherit less biases through partial data training. Given a full graph $\tilde{\mathcal{G}}=(\mathcal{V}, \mathcal{E}, \tilde{\textbf{X}})$ without the sensitive attribute, we have two partial versions of $\tilde{\mathcal{G}}$, i.e., $\tilde{\textbf{X}}$ and $\overline{\mathcal{G}}=(\mathcal{V}, \mathcal{E}, \overline{\textbf{X}})$. Then, $f_{tm}$, $f_{tg}$ generate node representations $H_{tm}$, $H_{tg}$ according to Eq. (\ref{eq:mexperts_forward})-(\ref{eq:gexperts_forward}). Taking $H_{tm}$, $H_{tg}$ as input, the classifier, e.g., a linear classification layer, is employed to predict node labels $\hat{y}_m$, $\hat{y}_g$. Based on the predicted node labels and the ground truth, BCE function is utilized as the loss function to optimize weights of two fairness experts, respectively. (2) Projector training ensures the projector generates informative node representations through a contrastive objective and a trained GNN $f_{cg}$. Here, $f_{cg}$ has the same network structure as our GNN student but is directly trained on the full graph $\tilde{\mathcal{G}}$. After fairness experts training, we fix the model parameters of $f_{tm}$, $f_{tg}$ and generate $H_{tm}$, $H_{tg}$. As shown in Eq. (\ref{eq:t_forward}), the projector takes the concatenation of $H_{tm}$, $H_{tg}$ as input and generates the node representations $H_{cg}^{'}$. This process can be summarized as $H_{cg}^{'}=f_{tp}(f_{tm}(\tilde{\textbf{X}}) \oplus f_{tg}(\overline{\mathcal{G}}))$. With node representations $H_{cg}=f_{cg}(\tilde{\mathcal{G}})$ generated by $f_{cg}$ as the ground truth, we optimize the projector $f_{tp}$ by minimizing the contrastive objective. This maximizes the similarity between the representation of the same node in $H_{cg}^{'}$ and $H_{cg}$, which means the synthetic model learns the informative node representations like $f_{cg}$. 

For the contrastive objective, we follow the setting in Zhu et al.~\cite{zhu2020deep}. $h_i$ and $h_i^{'}$ denote the node representation of node $i$ in $H_{cg}$ and $H_{cg}^{'}$. $(h_i, h_i^{'})$ can be considered as a positive pair and the remaining ones are all negative pairs. Given a similarity function $sim(\cdot)$ to computing node representations similarity, the contrastive objective for any positive pair $(h_i, h_i^{'})$ can be defined as:
\begin{equation}
\label{eq:contrastive_objective}
l(h_i, h_i^{'})=-\log(\frac{e^{sim(h_i, h_i^{'})/\tau}}{\sum_{j=1}^n e^{sim(h_i, h_j^{'})/\tau}+\sum_{j=1,i \neq j}^n e^{sim(h_i, h_j)/\tau}}),
\end{equation}
where $(h_i, h_j^{'})$ and $(h_i, h_j)$ are the negative pairs. $\tau$ is a scalar temperature parameter. We consider the similarity function $sim(\cdot)$ as $sim(h_i, h_i^{'})=\mathcal{D}(f(h_i),f(h_i^{'}))$, where $\mathcal{D}$ is the consine similarity and $f(\cdot)$ is a projector. We compute the contrastive objective over all positive pairs $(h_i, h_i^{'})$ to obtain the overall contrastive objective:
\begin{equation}
\label{eq:overall_contrastive_objective}
L_{con}=\frac{1}{2n} \sum_{i=1}^n [l(h_i, h_i^{'})+l(h_i^{'}, h_i)].
\end{equation}

With Eq. (\ref{eq:overall_contrastive_objective}) as the objective function, the trained synthetic teacher can generate fair and informative node representations for guiding the student model. As shown in Theorem~\ref{thm:informative}, we find that minimizing our contrastive objective is equivalent to maximizing the mutual information between $H_{cg}^{'}$ and the original graph $\tilde{\mathcal{G}}$. This proves that the node representation generated by $f_t$ is informative. As a result, FairGKD learns fair and informative GNN students with a such teacher model. 

\begin{theorem}
\label{thm:informative}
Let $H_{cg}$, $H_{cg}^{'}$ denote node representations generated by the trained GNN $f_{cg}$ and the synthetic teacher $f_{t}$, respectively. Given an original graph $\tilde{\mathcal{G}}$, the contrastive objective is a lower bound of mutual information between the node representations $H_{cg}^{'}$ generated by $f_{t}$ and the original graph $\tilde{\mathcal{G}}$:
\begin{equation}
\label{eq:theorem}
-L_{con} \leq I(H_{cg}^{'};\tilde{\mathcal{G}}).
\end{equation}
\end{theorem}

The proof of Theorem~\ref{thm:informative} is shown in Appendix~\ref{apd:proof}.


\subsection{Learning Fair GNNs Student}
With a trained synthetic model $f_{t}$ as the teacher model, FairGKD trains a fair GNN student $f_{s}$. Following the knowledge distillation paradigm \cite{hinton2015distilling}, the GNN student learns to mimic the output of the synthetic teacher. Assuming that $f_{s}$ is a $k$-layer model with a single linear layer as the classifier, i.e., the first $k-1$ layer constitutes the GNN backbone, while the $k$-th layer serves as the classifier. Given a full graph $\tilde{\mathcal{G}}=(\mathcal{V}, \mathcal{E}, \tilde{\textbf{X}})$, $f_{s}$ generates node representations $\Hat{H}=f_s^{(k-1)}(\tilde{\mathcal{G}})$ in ($k$-1)-th layer and predicts node labels $\hat{y}=f_s^{(k)}(\Hat{H})$ in $k$-th layer. With two partial versions of $\tilde{\mathcal{G}}$ as input, the trained $f_{t}$ distills the knowledge $H=f_t(\tilde{\textbf{X}}, \overline{\mathcal{G}})$ as the additional supervised information in soft loss. Node labels ground truth $Y$ is regarded as the ground truth in hard loss to make $f_{s}$ maintain high accuracy in the node classification task. Specifically, the objective function $L$ of training $f_{s}$ can be defined as:
\begin{equation}
\label{eq:loss_function}
    L=\alpha L_{c} + \beta L_{kd},
\end{equation}
where $L_{c}$ is the hard loss, which aims to maintain utility on the node classification task. We refer to $L_{c}$ as the binary cross-entropy function. $L_{kd}$ is the soft loss to optimize the GNN backbone for generating node representations similar to the output of $f_{t}$, which ensures that FairGKD learns fair GNNs. Similar to the optimization of the projector $f_{tp}$, we refer to $L_{kd}$ as the contrastive objective shown in Eq. (\ref{eq:contrastive_objective}), (\ref{eq:overall_contrastive_objective}) during the knowledge distillation process. $\alpha$ and $\beta$, which are adaptive coefficients for balancing the influence of these two loss terms, are calculated by the adaptive algorithm shown in Section \ref{subsec:ada_opt}.    

\subsection{Adaptive Optimization}
\label{subsec:ada_opt}
Due to the multi-loss of FairGKD, it can be challenging to achieve a balance between different loss terms, which indicates a trade-off between utility and fairness. To address this issue, an adaptive algorithm is used to calculate two adaptive coefficients (i.e., $\alpha$ and $\beta$) which balance the influence of the two loss terms. Inspired by the adaptive normalization loss in MTARD~\cite{zhao2022enhanced}, we use this algorithm to balance $L_{c}$ and $L_{kd}$ instead of loss terms in multi-teacher knowledge distillation. The core idea behind this algorithm is to amplify the impact of the disadvantage term (i.e., loss term with slower reduction) on the overall loss function. In this regard, the disadvantage term will be assigned a larger coefficient and then contribute more to the overall model update. Specifically, let $L(t)$ denote the loss at epoch $t$, Eq. (\ref{eq:loss_function}) can be formulated
as:
\begin{equation}
\label{eq:loss_function_t}
    L(t)=\alpha(t) L_{c}(t) + \beta(t) L_{kd}(t).
\end{equation}

We compute $\alpha(t)$ and $\beta(t)$ at each epoch $t$ through the loss decrease relative to the initial epoch, i.e., $t=0$. Here, we utilize the relative loss $\tilde{L}(t)=L(t)/L(0)$ to measure the loss decrease of loss $L$. A small value of $\tilde{L}(t)$ means a better optimization for the training model. Based on $\tilde{L}(t)$, $\alpha(t)$ and $\beta(t)$ can be computed by the following formulation:
\begin{align}
\label{eq:ada_coefficient}
    \alpha(t) &= \frac{lr[\tilde{L}_{c}(t)]^{\gamma}}{[\tilde{L}_{c}(t)]^{\gamma}+[\tilde{L}_{kd}(t)]^{\gamma}} +(1-lr)\alpha(t-1), \\
    \beta(t) &= 1-\alpha(t),
\end{align}
where $lr$ is the learning rate of the hard loss term, $\gamma$ is a hyperparameter to enhance the disadvantaged loss which represents the loss term with a lower loss decrease. 

In our scenarios, this algorithm suppresses the significantly decreasing loss term by assigning a smaller coefficient. Conversely, a bigger coefficient will promote the loss term that is slightly decreased. Meanwhile, $\alpha$ and $\beta$ are dynamically changing with model optimization. Thus, the GNN student is optimized by balancing two loss terms, resulting in the trade-off between utility and fairness. Additionally, we give a brief complexity analysis for our proposed method FairGKD in Appendix~\ref{apd:complexity}.


\section{Experiments}
\label{sec:exeps}
In this section, we conduct experiments on three datasets, namely, $Recidivism$~\cite{agarwal2021towards}, $Pokec$-$z$, $Pokec$-$n$~\cite{takac2012data}. The statistical information of these datasets is shown in Table~\ref{tab:statistic}. Limited by space, we present the detailed experimental settings in Appendix~\ref{apd:settings}. 

\begin{table}[!t]
    \centering
    \caption{Statistic information of three datasets.}
    \resizebox{0.95\linewidth}{!}{
    \begin{tabular}{lccc}
        \toprule
        \textbf{Dataset}  & \textbf{\textit{Recidivism}} & \textbf{\textit{Pokec-z}} & \textbf{\textit{Pokec-n}} \\
        \midrule
        \#Nodes            & 18,876           & 67,796         & 66,569  \\
        \#Edges            & 321,308          & 617,958        & 583,616   \\
        \#Attributes       & 18               & 277            & 266   \\
        Sens.               & Race                & Region         & Region  \\
        Labels              & Bail Prediction     & Working Field  & Working Field \\
        \bottomrule
    \end{tabular}}
    \label{tab:statistic}
\end{table}

\begin{table*}[!ht]
    \centering
    \caption{Comparison results of FairGKD against baseline methods. FairGKD$\backslash$S is a variant of FairGKD with the same network structure and is trained on graph data without the sensitive attribute. The best results for each backbone GNN are in bold.}
    \renewcommand\arraystretch{1.1}
    \setlength{\tabcolsep}{0.5mm}{
    \begin{tabular}{cc|ccc|ccc|ccc}
    \toprule
    \multicolumn{1}{l}{\multirow{2}{*}{}} & \multirow{2}{*}{\textbf{Methods}} & \multicolumn{3}{c}{\textbf{\textit{Recidivism}}}                                & \multicolumn{3}{c}{\textbf{\textit{Pokec-z}}}                                & \multicolumn{3}{c}{\textbf{\textit{Pokec-n}}}      \\ \cmidrule(l){3-11}
    \multicolumn{1}{l}{}                  &           & ACC ($\uparrow$)          & $\Delta_{DP}$ ($\downarrow$)  & $\Delta_{EO}$ ($\downarrow$)           & ACC ($\uparrow$)          & $\Delta_{DP}$ ($\downarrow$)  & $\Delta_{EO}$ ($\downarrow$)   & ACC ($\uparrow$)          & $\Delta_{DP}$ ($\downarrow$)  & $\Delta_{EO}$ ($\downarrow$) \\ \hline
    \multirow{6}{*}{\textbf{GCN}} & Vanilla              & 84.23 ± 0.71          & 7.54 ± 0.59          & 5.20 ± 0.52          & \textbf{69.74 ± 0.27} & 5.02 ± 0.56           & 4.15 ± 0.81          & 68.75 ± 0.36         &  1.58 ± 0.57          & 2.56 ± 1.07 \\
                                & FairGNN              & 82.93 ± 1.17          & 6.65 ± 0.80          & 4.50 ± 0.90 & 68.96 ± 1.52         & 7.95 ± 1.53           & 6.29 ± 2.00          & 67.53 ± 1.75         &  2.36 ± 1.25          & 2.87 ± 1.05 \\ 
                                & FairVGNN             & 85.92 ± 0.57          & 6.00 ± 0.75          & 4.69 ± 1.19          & 68.30 ± 0.94         & \textbf{3.51 ± 3.01}  & 3.67 ± 1.88          & \textbf{69.25 ± 0.29}&  6.91 ± 1.74          & 8.93 ± 1.85 \\ 
                                & FDKD                 & 84.12 ± 0.52          & 7.33 ± 0.77          & 6.40 ± 1.14          & 69.71 ± 0.26         & 4.36 ± 0.68           & 3.25 ± 0.66          & 68.70 ± 0.53         &  1.03 ± 0.58          & 1.92 ± 0.83 \\ \cmidrule{2-11}
                                & FairGKD              & \textbf{86.10 ± 1.09}          & 7.12 ± 1.15          & 5.86 ± 1.30          & 69.21 ± 0.41         & 4.16 ± 0.80           & \textbf{3.15 ± 0.85} & 67.85 ± 0.30         &  \textbf{0.87 ± 0.50} & \textbf{1.25 ± 1.04} \\
                                & FairGKD$\backslash$S & 86.09 ± 1.05 & \textbf{5.98 ± 0.47} & \textbf{4.47 ± 0.72}          & 69.42 ± 0.37         & 7.72 ± 0.90           & 5.90 ± 0.99          & 68.04 ± 0.54         &   2.46 ± 0.86          & 3.62 ± 0.94 \\ 
                                \cmidrule(l){1-11}
    \multirow{6}{*}{\textbf{GIN}} & Vanilla              & 72.53 ± 4.95          & 8.96 ± 3.13          & 6.62 ± 2.21          & 68.37 ± 0.55         & 4.29 ± 1.84           & 4.65 ± 1.85          & \textbf{68.48 ± 0.40}&  2.75 ± 1.45          & 4.81 ± 2.84 \\ 
                                & FairGNN              & 78.34 ± 1.56          & 8.93 ± 1.63          & 6.65 ± 1.77          & 68.48 ± 0.79         & 3.96 ± 1.47           & 5.22 ± 1.51          & 67.50 ± 1.11         &  2.25 ± 1.33          & 2.68 ± 1.59 \\ 
                                & FairVGNN             & \textbf{80.76 ± 1.22}          & 6.95 ± 0.41          & 6.97 ± 1.18          & 68.35 ± 0.71         & 1.93 ± 1.23  & 2.71 ± 1.20 & 67.55 ± 1.33         &  6.13 ± 1.59          & 7.00 ± 1.80 \\ 
                                & FDKD                 & 61.29 ± 6.17          & 5.95 ± 2.90          & 4.93 ± 3.54          & 68.40 ± 0.50         & 3.39 ± 1.19           & 4.16 ± 1.52          & 66.56 ± 1.57         & 1.60 ± 1.04 & 2.90 ± 2.27 \\ \cmidrule{2-11}
                                & FairGKD              & 78.09 ± 4.53          & 8.37 ± 1.52          & 5.57 ± 1.63          & 67.63 ± 2.10         & \textbf{1.60 ± 1.07}           & \textbf{2.26 ± 1.66}          & 67.90 ± 1.09         &  \textbf{1.59 ± 0.99}          & \textbf{1.81 ± 1.59} \\
                                & FairGKD$\backslash$S & 79.98 ± 2.45 & \textbf{5.89 ± 1.56} & \textbf{4.53 ± 0.96} & \textbf{68.87 ± 0.35} & 4.78 ± 1.03           & 5.44 ± 1.38          & \textbf{68.48 ± 0.63}         &   2.10 ± 0.81          & 4.57 ± 1.32 \\   
        \hline
    \end{tabular}}
\small
\label{tab:comparison}
\end{table*}

\subsection{Comparison with Baseline Methods}

With two different GNNs (GCN, GIN) as backbones, we compare FairGKD with three baseline methods on the node classification task. Specifically, we implement two FairGKD variants with the same network structure, i.e., {\itshape FairGKD} and {\itshape FairGKD$\backslash$S}, which represent training on graph-structured data with and without the sensitive attribute. 
Note that we denote the proposed method in Section~\ref{sec:method} as FairGKD$\backslash$S for convenience. Table \ref{tab:comparison} presents the utility and fairness performance of two FairGKD variants and all baseline methods.

The following observations can be made from Table \ref{tab:comparison}: (1) the proposed method FairGKD improves fairness while maintaining or even improving utility across all three datasets, highlighting the effectiveness of FairGKD in learning fair GNNs. 
(2) FairGKD achieves state-of-the-art performance in most cases. {\itshape While it is not the best-performing method in some cases, FairGKD presents a small performance gap with the best result.} This implies the superior performance of FairGKD on both fairness and utility. (3) FairGKD can improve the fairness of GNNs even without sensitive attribute information, and in some cases outperforms all baseline methods. This observation implies that FairGKD is a demographic-agnostic fairness method.
 %
 
\begin{figure}[!t]
  \centering
  \includegraphics[width=0.95\linewidth]{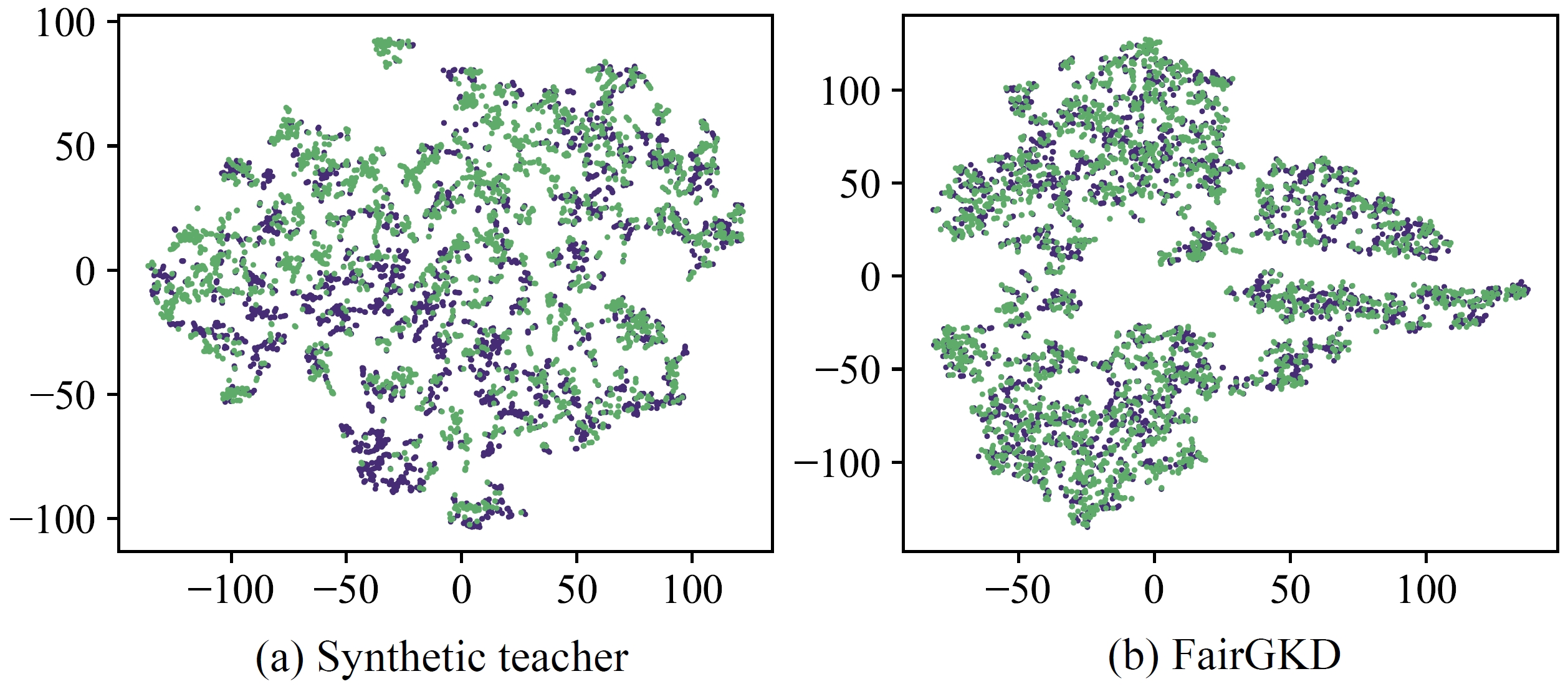}
  \caption{Visualizations of node representation learned on \textit{Recidivism} dataset. Dot colors represent two demographic groups with different sensitive attributes (race).}
  \label{fig:visual_sens}
\end{figure}

\begin{figure}[!t]
  \centering
  \includegraphics[width=0.95\linewidth]{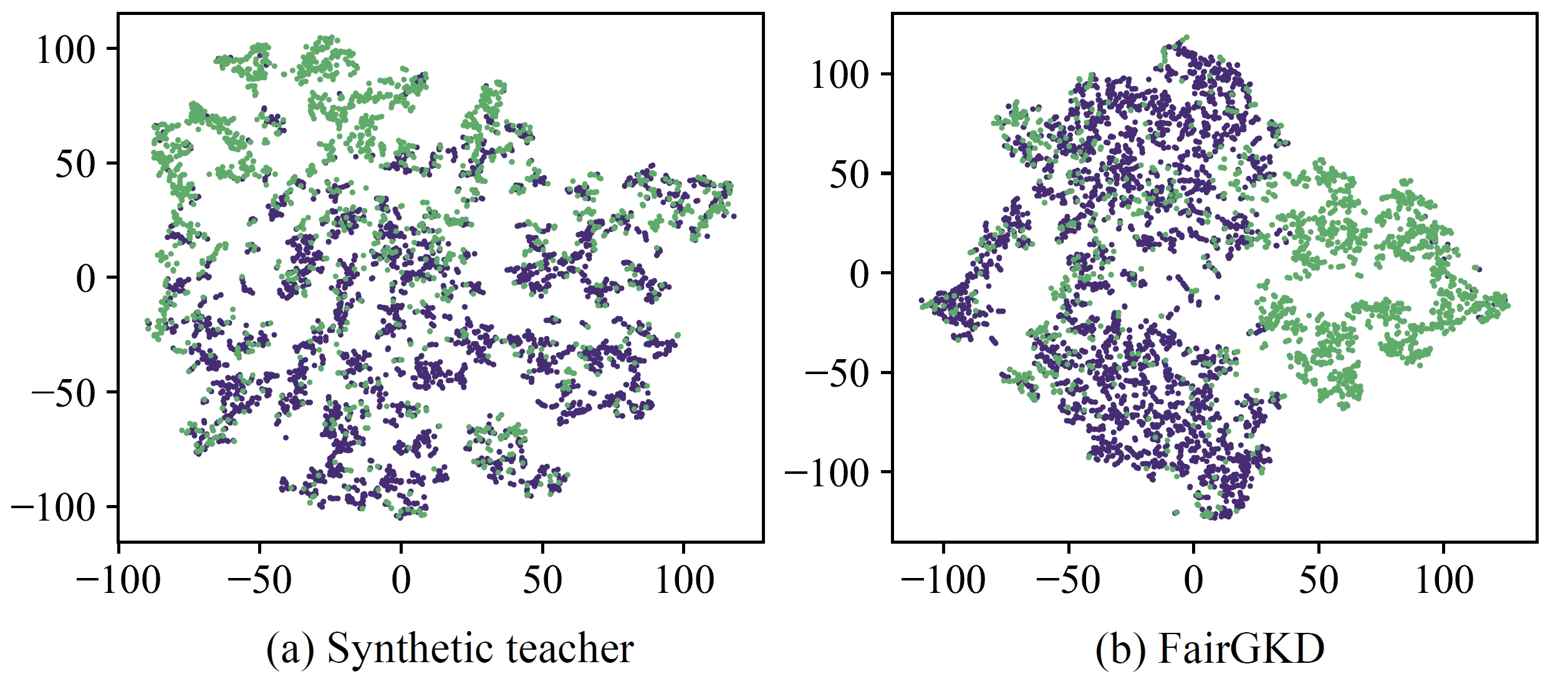}
  \caption{Visualizations of node representation learned on \textit{Recidivism} dataset. Dot colors represent the label (bail vs. no bail) of nodes.}
  \label{fig:visual_labels}
\end{figure}

 We also find that FairGKD$\backslash$S does not improve fairness on  $Pokec$-$z$ and $Pokec$-$n$ dataset, but rather performs worse than {\itshape Vanilla}. However, comparing FairGKD$\backslash$S with {\itshape Vanilla$\backslash$S} trained on data without the sensitive attribute, we observe  FairGKD$\backslash$S improves the fairness of GNNs. One potential reason could be that non-sensitive attributes, which are highly correlated to the sensitive attribute, leak sensitive information and are more prone to make the training GNN inherit bias. We also observe that FairGKD is effective for two commonly used GNNs. To further understand FairGKD, we visualize the node representation generated by the synthetic teacher and the trained GNN student of FairGKD on $Recidivism$ using t-SNE in Figures \ref{fig:visual_sens} and \ref{fig:visual_labels}. Figure \ref{fig:visual_sens} shows indistinguishable node representations with respect to the sensitive attribute, which demonstrates fair knowledge distilled by the synthetic teacher and fair student learned by FairGKD. Figure \ref{fig:visual_labels} shows clear clustering of node representations with respect to the node label, indicating the superior utility performance of FairGKD.

\subsection{Ablation Study}
We conduct ablations on datasets with the sensitive attribute. Specifically, we investigate how each of the following four components individually contributes to learning fair GNNs, i.e., fairness GNN expert, fairness MLP expert, the projector, and the adaptive optimization algorithm. We remove these four components separately and their results denoted by {\itshape FairGKD$\backslash$G}, {\itshape FairGKD$\backslash$M}, {\itshape FairGKD$\backslash$P}, and {\itshape FairGKD$\backslash$A}, respectively. For FairGKD$\backslash$G and FairGKD$\backslash$M, we regard the output of the remaining expert as the input of the projector. FairGKD$\backslash$P replaces the projector with {\itshape mean} operation. FairGKD$\backslash$A fixes $\alpha$ and $\beta$ in Eq. (\ref{eq:loss_function}) as 0.5. 

\begin{figure}[!t]
  \centering
  \includegraphics[width=\linewidth]{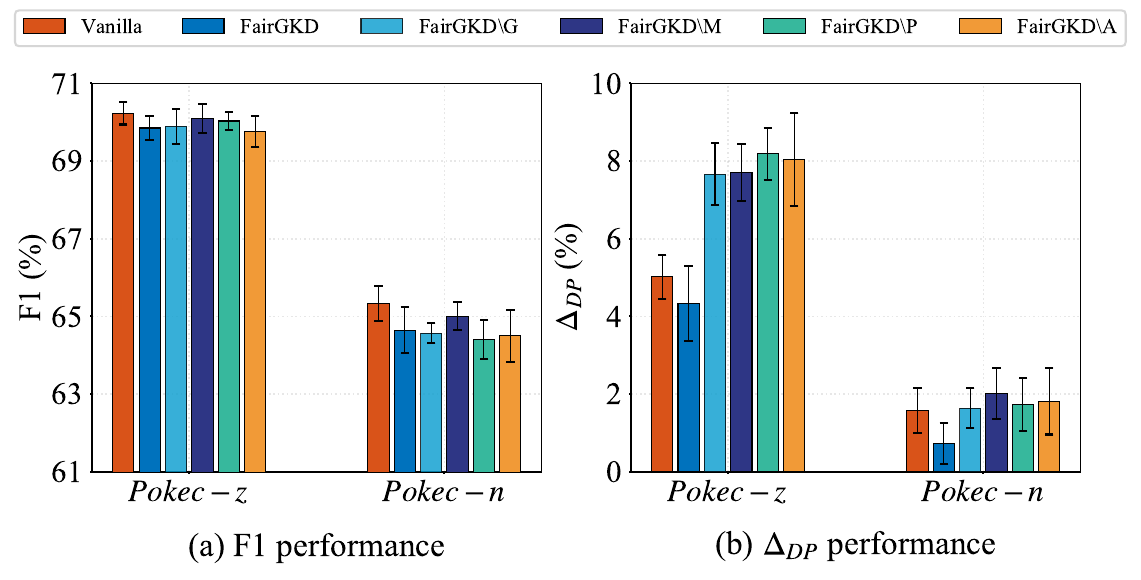}
  \caption{Ablation results w.r.t. the components of FairGKD.} 
  \label{fig:ablation}
\end{figure}

We show ablation results of FairGKD learning GCN student model on $Pokec$-$z$, and $Pokec$-$n$ datasets in Figure \ref{fig:ablation}. We observe that FairGKD is stable in utility performance even if lacks some components. This is due to the fact that FairGKD trains the GNN student with full data as input and the knowledge distilling from the synthetic teacher is informative. The removal of any components, especially the fairness GNN expert and the fairness MLP expert, weakens the fairness improvement of FairGKD. This observation indicates that the synthetic teacher and the adaptive algorithm for loss coefficients are necessary for learning a fair GNN. Moreover, results of FairGKD$\backslash$A show higher $\Delta_{DP}$ value and standard deviations, which means worse fairness and more unstable performance. Thus, the adaptive algorithm for calculating loss coefficients is effective in achieving the trade-off between fairness and utility. Overall, the results show that all components of FairGKD are beneficial for the model fairness performance, while the adaptive algorithm also ensures the model fairness performance remains stable.


\subsection{Various Sensitive Attributes Evaluation} 
Due to the partial data training to mitigate higher-level biases, FairGKD can be employed in fairness scenarios involving various sensitive attributes within a single training session. To verify this, we conduct FairGKD with GCN as the student and evaluate the fairness performance of this student on different sensitive attributes. We study the fairness performance of FairGKD w.r.t. three sensitive attributes, i.e., region, gender, and age. Here, we set all sensitive attributes to be binary. We set age $\ge$ 25 as 1 and age $<$ 25 as 0. 

The results of various sensitive attributes are shown in Table \ref{tab:various}. We observe that a single implementation of FairGKD improves fairness under different sensitive attributes in most cases. This is due to the fact that FairGKD is not tailored for a specific sensitive attribute and mitigates higher-level biases. Due to such a design, FairGKD exhibits a weaker impact on fairness in sensitive attributes like gender. Moreover, we also find that {\itshape vanilla} outperforms FairGKD on utility performance by a small margin which can be ignored. 
Overall, experimental results demonstrate that FairGKD is a one-size-fits-all approach for improving fairness under different sensitive attributes. 

\begin{table}[!t]
\centering
\caption{Results of various sensitive attributes.} 
\renewcommand\arraystretch{1.1}
\resizebox{\linewidth}{!}{
\begin{tabular}{@{}cccccc@{}}
\toprule
    \multicolumn{2}{c}{\multirow{2}{*}{\textbf{Metrics}}} & \multicolumn{2}{c}{\textbf{\textit{Pokec-z}}} & \multicolumn{2}{c}{\textbf{\textit{Pokec-n}}} \\ \cmidrule(l){3-6} 
    \multicolumn{2}{l}{} & \multicolumn{1}{c}{Vanilla} & \multicolumn{1}{c}{FairGKD} & \multicolumn{1}{c}{Vanilla} & \multicolumn{1}{c}{FairGKD} \\ \midrule
    \multicolumn{2}{c}{F1}                 & \textbf{70.23 ± 0.29}               & 69.59 ± 0.36                & \textbf{65.33 ± 0.45}               & 64.47 ± 0.47               \\
    \multicolumn{2}{c}{ACC}                      & \textbf{69.74 ± 0.27}                & 69.21 ± 0.41                & \textbf{68.75 ± 0.36}               & 67.85 ± 0.30                 \\ \midrule
    \multirow{3}{*}{$\Delta_{DP}$}   & Region         & 5.02 ± 0.56                & \textbf{4.16 ± 0.80}                & 1.58 ± 0.57                & \textbf{0.87 ± 0.50}                 \\
                          & Gender   & 3.07 ± 0.51                & \textbf{2.56 ± 0.72}                & \textbf{6.03 ± 0.89}                 & 6.42 ± 0.45                \\
                          & Age      & 57.03 ± 0.62               & \textbf{55.41 ± 2.82}               & 62.41 ± 1.01               & \textbf{59.24 ± 2.08}               \\ \midrule
    \multirow{3}{*}{$\Delta_{EO}$} & Region & 4.15 ± 0.81                & \textbf{3.15 ± 0.85}                & 2.56 ± 1.07                & \textbf{1.25 ± 1.04}                \\
                          & Gender & 5.03 ± 0.37                & \textbf{5.01 ± 0.90}                 & 12.84 ± 0.99               & \textbf{12.43 ± 1.06}               \\
                          & Age            & 44.88 ± 0.53               & \textbf{43.52 ± 3.28}               & 51.53 ± 1.10                & \textbf{48.72 ± 2.04}               \\ 
\bottomrule
\end{tabular}
}
\label{tab:various}
\end{table}

\begin{figure}[t]
\centering
\subfigure[ACC performance]{
\includegraphics[width=0.45\columnwidth]{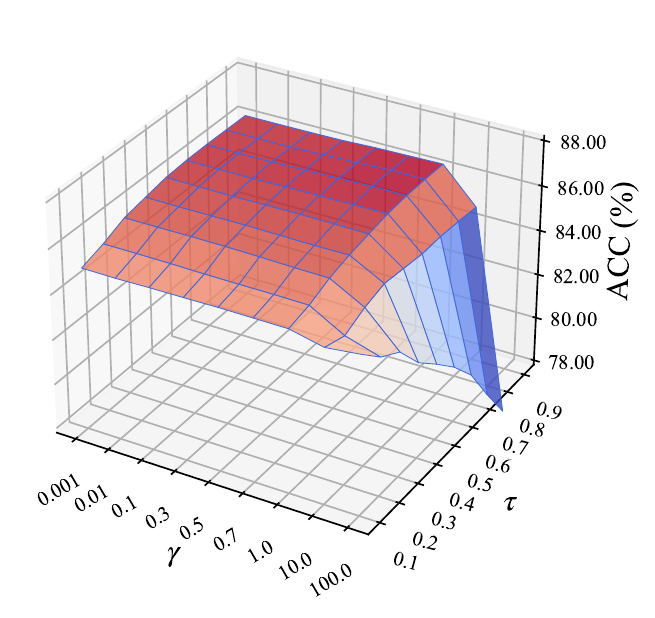}
}
\subfigure[$\Delta_{DP}$ performance]{
\includegraphics[width=0.45\columnwidth]{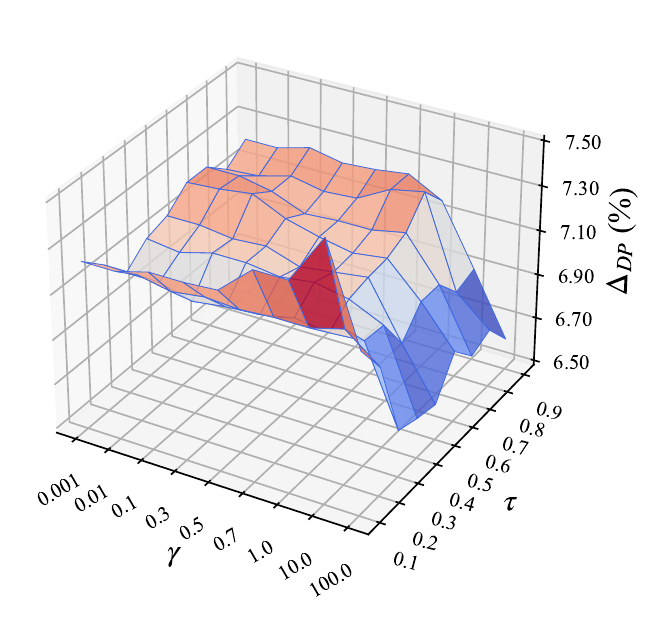}
}
\quad
\caption{Parameters sensitivity analysis on \textit{Recidivism}.}
\label{fig:paras_sens}
\end{figure}

\subsection{Parameters Sensitivity} 
We conduct hyperparameter studies of two parameters, i.e., $\tau$ and $\gamma$. Specifically, we only show the parameter sensitivity results on $Recidivism$ dataset due to similar observations on other datasets. Using GCN as the student model, we vary $\tau$ and $\gamma$ within the range of $\{$0.1, 
 0.2, 0.3, 0.4, 0.5, 0.6, 0.7, 0.8, 0.9$\}$ and $\{$0.001, 0.01, 0.1, 0.3, 0.5, 0.7, 1, 10, 100$\}$, respectively. As shown in Figure \ref{fig:paras_sens}, we make the following observations: (1) the overall performance of FairGKD remains smooth despite the wide range of variation in $\gamma$ and $\tau$. Specifically, the performance of FairGKD is stable in terms of utility as well as fairness when $\gamma$ varies within the range of 0.001 to 10 or $\tau$ varies within the range of 0.1 to 0.9. (2) When $\gamma \ge 10$, FairGKD further improves fairness but the utility performance decreases rapidly. Due to a larger $\gamma$ strengthening the disadvantaged loss, this observation indicates that the soft loss $L_{kd}$ is the disadvantaged one in training. In this regard, a larger $\gamma$ improves fairness but sacrifices utility. To better achieve the trade-off between utility and fairness, a suitable $\gamma$ is necessary. (3) When $\tau \leq 0.1$, the fairness and utility performance of FairGKD both decrease. This suggests that the GNN student is not accurately replicating the output of the synthetic teacher. Overall, FairGKD remains stable with respect to the wide range variation of $\gamma$ and $\tau$. A suitable $\gamma$, e.g., $\gamma < 10$, is beneficial for the trade-off between utility and fairness. 


\section{Conclusion}
In this paper, we propose FairGKD, a simple yet effective method to learn fair GNNs on graph-structured data through knowledge distillation. To our best knowledge, FairGKD is the first effort to improve GNN fairness on graph-structured data without demographic information. The key insight behind our approach is to guide the GNN student training through the fair synthetic teacher, which mitigates higher-level biases via partial data training, i.e., training models on only node attributes or graph topology. As a result, FairGKD leverages the benefits of knowledge distillation and partial data training, improving fairness while eliminating the dependence on demographics. Experiments on three real-world datasets demonstrate that FairGKD outperforms strong baselines in most cases. Furthermore, we demonstrate that FairGKD is effective in learning fair GNNs even when demographic information is absent or in multiple sensitive attribute scenarios. 


\section{Acknowledgements}
The research is supported by the National Key R$\&$D Program of China under grant No. 2022YFF0902500, the Key-Area Research and Development Program of Shandong Province (2021CXGC010108), the Guangdong Basic and Applied Basic Research Foundation, China (No. 2023A1515011050). 

\bibliographystyle{ACM-Reference-Format}
\bibliography{main}

\clearpage
\appendix
\section{Proof of theorem 1}
\label{apd:proof}

\addtocounter{theorem}{-1}
\theoremstyle{plain}
\begin{theorem}[Restated]\label{thm:informative}
Let $H_{cg}$, $H_{cg}^{'}$ denote node embedding generated by the trained GNN $f_{cg}$ and the synthetic teacher $f_{t}$, respectively. Given an original graph $\mathcal{G}$, the contrastive objective is a lower bound of mutual information between the node embedding $H_{cg}^{'}$ generated by $f_{t}$ and the original graph $\mathcal{G}$:
\begin{equation}
\label{eq:theorem}
-L_{con} \leq I(H_{cg}^{'};\mathcal{G}).
\end{equation}
\end{theorem}

Here is the proof of Theorem~\ref{thm:informative}:

\begin{proof}
Based on the data processing inequality, given three random variables \textbf{X}, \textbf{Y}, \textbf{Z} satisfying a Markov chain $\textbf{X}\rightarrow\textbf{Y}\rightarrow \textbf{Z}$, we have the inequality $I(\textbf{X}; \textbf{Z}) \leq I(\textbf{X}; \textbf{Y})$. In FairGKD, $H_{cg}$ and $H_{cg}^{'}$ are generated by $f_{cg}$ and $f_{t}$ using the original graph $\tilde{\mathcal{G}}$ as input. This gives rise to a Markov chain $H_{cg}^{'} \leftarrow \tilde{\mathcal{G}} \rightarrow H_{cg}$. $H_{cg}^{'}$ and $H_{cg}$ are conditionally independent after given $\tilde{\mathcal{G}}$, so this Markov chain is equivalent to $H_{cg}^{'} \rightarrow \tilde{\mathcal{G}} \rightarrow H_{cg}$. Thus, we obtain the following inequality: 
\begin{equation}
\label{eq:proof_1}
I(H_{cg}^{'}; H_{cg}) \leq I(H_{cg}^{'}; \tilde{\mathcal{G}}).
\end{equation} 

According to the proof in \cite{ling2023learning,zhu2020deep}, the lower bound of the true mutual information between $H_{cg}$ and $H_{cg}^{'}$ is the contrastive objective $-L_{con}$, which can be summarized as:
\begin{equation}
\label{eq:proof_2}
-L_{con} \leq I(H_{cg}^{'};H_{cg}).
\end{equation}

Combining Eq. (\ref{eq:proof_1}) and (\ref{eq:proof_2}), the following inequality can be summarized:
\begin{equation}
\label{eq:proof_3}
-L_{con} \leq I(H_{cg}^{'};H_{cg}) \leq I(H_{cg}^{'}; \tilde{\mathcal{G}}).
\end{equation}

This completes the proof of Theorem \ref{thm:informative}, and we have shown that $-L_{con} \leq I(H_{cg}^{'}; \tilde{\mathcal{G}})$.
\end{proof}

\section{Complexity Analysis}
\label{apd:complexity}
We briefly discuss the space and time complexity of our proposed FairGKD. For space complexity, FairGKD computes the pairwise similarity in the contrastive objective to train fair GNNs. Thus, the space complexity of FairGKD, which is related to the number of nodes in graph $\tilde{\mathcal{G}}=(\mathcal{V}, \mathcal{E}, \tilde{\textbf{X}})$, is $\mathcal{O}(\lvert \mathcal{V} \rvert^2)$. Fortunately, we can use the graph sampling-based training method or replace the contrastive objective with mean square error (MSE) to further reduce the space complexity. For time complexity, FairGKD conducts the forward of
the synthetic teacher once to obtain soft targets and the rest of the training process is the same as the normal training of GNNs. Thus, the time complexity of FairGKD is $\mathcal{O}(\lvert \mathcal{V} \rvert + \lvert \mathcal{E} \rvert)$.

\section{Algorithm}
\label{apd:algorithm}
To help better understand FairGKD, we present the training algorithm of FairGKD in Algorithm~\ref{alg:train_algo}. Note that the trained GNN $f_{cg}$ has the same network structure as the GNN student but is directly trained on the full graph $\tilde{\mathcal{G}}$.

\begin{algorithm}[]
\caption{Training Algorithm of FairGKD}
\label{alg:train_algo}
    \begin{flushleft}
        \textbf{Input}: $\tilde{\mathcal{G}}=(\mathcal{V}, \mathcal{E}, \tilde{\textbf{X}})$, node labels $y$, a $k$-layer GNN student $f_s$, a synthetic teacher $f_t=\{f_{tm}, f_{tg}, f_{tp}\}$, a trained GNN $f_{cg}$, and hyperparameters $\tau$, $\gamma$, $lr$. \\
        \textbf{Output}: Trained GNN student model with parameters $\theta_{f_{s}}$.
    \end{flushleft}
    \begin{algorithmic}[1] 
        \STATE // Fairness experts training
        \STATE Train $f_{tm}$, $f_{tg}$ with the binary cross-entropy function as the loss function;
        \STATE // Projector training
        \STATE $\overline{\mathcal{G}}$, $\tilde{\textbf{X}}$ $\leftarrow$ $\tilde{\mathcal{G}}$;
        \WHILE{$not~converged$}
        \STATE // Frozen $f_{tm}$, $f_{tg}$
        \STATE $H_{tm}$ $\leftarrow$ $f_{tm}(\tilde{\textbf{X}})$, $H_{tg}$ $\leftarrow$ $f_{tg}(\overline{\mathcal{G}})$;
        \STATE $H_{cg}^{'} \leftarrow f_{tp}(H_{tm} \oplus H_{tg})$, $H_{cg} \leftarrow f_{cg}(\tilde{\mathcal{G}})$;
        \STATE Calculate $L_{con}$ according to Eq.(\ref{eq:contrastive_objective}) and (\ref{eq:overall_contrastive_objective});
        \STATE Update $\theta_{f_{tp}}$ by gradient descent;
        \ENDWHILE
        \STATE // Fair GNN student training (knowledge distillation)
        \STATE $H \leftarrow f_{tp}(f_{tm}(\tilde{\textbf{X}}) \oplus f_{tg}(\overline{\mathcal{G}}))$;
        \WHILE{$not~converged$}
        \STATE $\hat{y} \leftarrow f_s^{(k)}(\Hat{H})$, $\Hat{H} \leftarrow f_s^{(k-1)}(\tilde{\mathcal{G}})$;
        \STATE Calculate $L_{c}(t)$, $L_{kd}(t)$;
        \STATE Calculate $\alpha(t)$, $\beta(t)$ according to Eq.(\ref{eq:ada_coefficient}) and (15);
        \STATE Calculate loss function  $L \leftarrow \alpha(t)L_{c}(t)+\beta(t)L_{kd}(t)$;
        \STATE Update $\theta_{f_{s}}$ by gradient descent;
        \ENDWHILE
        \STATE \textbf{return} $\theta_{f_{s}}$;
    \end{algorithmic}
\end{algorithm}

\section{Experimental Settings}
\label{apd:settings}

\subsection{Datasets}
\label{apd:dataset}
We conduct node classification experiments on three commonly used datasets, namely, $Recidivism$~\cite{agarwal2021towards}, $Pokec$-$z$, $Pokec$-$n$~\cite{takac2012data}. To maintain consistency with previous research~\cite{dai2021say,wang2022improving,dong2022edits}, the sensitive attribute and node labels of the above-mentioned datasets are binary. We present a brief description of these three datasets as follows:
\begin{itemize}
\item \textbf{Recidivism} $Recidivism$~\cite{agarwal2021towards} is a defendants dataset including those defendants released on bail during 1990-2009 in U.S states. Nodes represent defendants and are connected based on the similarity of past criminal records and demographics. Considering ``race'' as the sensitive attribute, the task is to classify defendants into bail vs. no bail, i.e., predicting whether defendants will commit a crime after being released. 
\item \textbf{Pokec} $Pokec$-$z$ and $Pokec$-$n$~\cite{takac2012data} are derived from a popular social network application in Slovakia, which shows different data in two different provinces. A node denotes a user with features such as gender, age, interest, etc. A connection represents the friendship information between nodes. Regarding ``region'' as the sensitive attribute, we predict the working field of the users.
\end{itemize}

\subsection{Evaluation Metrics}
\label{apd:metrics}
We regard the F1 score and the accuracy as utility evaluation metrics. For fairness performance, we use two group fairness metrics (i.e., $\Delta_{DP}$ and $\Delta_{EO}$) shown in Section \ref{sec:metrics}. A smaller fairness metric indicates a fairer model decision.

\subsection{Baseline Methods}
\label{apd:baseline}
We compare FairGKD with the following baseline methods, including FairGNN~\cite{dai2021say}, FairVGNN~\cite{wang2022improving}, and FDKD~\cite{chai2022fairness}:
\begin{itemize}
    \item \textbf{FairGNN.} FairGNN focuses on the group fairness of graph-structured data with limited sensitive attribute information. Based on adversarial learning, FairGNN aims to train a fair GNN classifier to generate results for which the discriminator cannot distinguish sensitive attributes.   
    \item \textbf{FairVGNN.} FairVGNN improves the fairness of GNNs by mitigating sensitive attribute leakage. Specifically, it aims to mask features and clamp encoder weights highly relevant to the sensitive attribute.
    
    \item \textbf{FDKD.} FDKD, which is inspired by the fairness impact of label smoothing, focuses on fairness without demographics on IID data via soft labels from an overfitted teacher model. Both FDKD and FairGKD follow knowledge distillation, but they are quite different. To employ FDKD in graph-structured data, we replace the original backbone (ResNet~\cite{he2016deep}) in FDKD with the commonly used GNN backbones, e.g., GCN~\cite{kipf2016semi} and GIN~\cite{xu2018powerful}.
\end{itemize}

\subsection{Implemental Details}
\label{apd:imple_detail}
We conduct all experiments 10 times and reported average results. For FairGKD, we fix the learning rate as 1.0 for all datasets. We set $\tau=\{0.5, 0.9\}$ and $\gamma=\{0.1, 0.001\}$ for the $Recidivism$, $Pokec$-$n$ dataset. For $Pokec$-$z$ dataset, we set $\tau=\{0.5, 0.9\}$ and $\gamma=\{0.001, 0.001\}$ for GCN, GIN backbone. For FairGNN and FairVGNN, we train models while all hyperparameters follow the author's suggestions. For FDKD, we tune the trade-off hyperparameter to find the best model performance in utility and fairness. For all models, we utilize the Adam optimizer with a learning rate of $1 \times 10^{-3}$  and a weight decay scheme for 1000 epochs, where the weight decay is set to $1 \times 10^{-5}$. 

Since baseline methods (i.e., FairGNN, FairVGNN) cannot be used in a scenario without sensitive attributes, all FairGKD implementations in experiments use the sensitive attribute for a fair comparison.

We use two commonly used GNNs (i.e., GCN, and GIN) as the GNN students in FairGKD. For GCN and GIN, a 1-layer GCN and GIN convolution, respectively, serve as the backbone, with a single linear layer positioned on top of the backbone as the classifier. For the synthetic teacher, FairGKD uses a 2-layer MLP as the fairness MLP expert, a 1-layer GCN as the fairness GNN expert, and a 3-layer MLP as the projector. Meanwhile, the hidden dimensions of all GNNs and MLPs for three datasets are 16. Moreover, we conduct all experiments on one NVIDIA TITAN RTX GPU with 24GB memory. All models are implemented with PyTorch.

\end{document}